\documentclass[10pt,twocolumn,letterpaper]{article}

\usepackage{cvpr}      

\usepackage{graphicx}
\usepackage{amsmath}
\usepackage{amssymb}
\usepackage{booktabs}

\usepackage[pagebackref,breaklinks,colorlinks]{hyperref}

\usepackage[capitalize]{cleveref}
\crefname{section}{Sec.}{Secs.}
\Crefname{section}{Section}{Sections}
\Crefname{table}{Table}{Tables}
\crefname{table}{Tab.}{Tabs.}

\def\cvprPaperID{*****} 
\def\confName{CVPR}
\def\confYear{2024}

\usepackage{array}
\newcolumntype{P}[1]{>{\centering\arraybackslash}p{#1}}

\begin{document}

\title{PBVS 2024 Solution: Self-Supervised Learning and Sampling Strategies
\\ 
for SAR Classification in Extreme Long-Tail Distribution}

\author{Yuhyun Kim\thanks{equal contribution} \quad Minwoo Kim\footnotemark[1] \quad Hyobin Park\footnotemark[1] \quad Jinwook Jung\footnotemark[1]  \quad Dong-Geol Choi\thanks{corresponding author} \\
Hanbat National University \\
{\tt\small \{yuhyun.kim, minwoo.kim, hyobin.park, jinwook.jung\}@edu.hanbat.ac.kr} \\
{\tt\small dgchoi@hanbat.ac.kr}
}
\maketitle

\begin{abstract}
The Multimodal Learning Workshop (PBVS 2024) aims to improve the performance of automatic target recognition (ATR) systems by leveraging both Synthetic Aperture Radar (SAR) data, which is difficult to interpret but remains unaffected by weather conditions and visible light, and Electro-Optical (EO) data for simultaneous learning. The subtask, known as the Multi-modal Aerial View Imagery Challenge - Classification, focuses on predicting the class label of a low-resolution aerial image based on a set of SAR-EO image pairs and their respective class labels. The provided dataset consists of SAR-EO pairs, characterized by a severe long-tail distribution with over a 1000-fold difference between the largest and smallest classes, making typical long-tail methods difficult to apply. Additionally, the domain disparity between the SAR and EO datasets complicates the effectiveness of standard multimodal methods.

To address these significant challenges, we propose a two-stage learning approach that utilizes self-supervised techniques, combined with multimodal learning and inference through SAR-to-EO translation for effective EO utilization. In the final testing phase of the PBVS 2024 Multi-modal Aerial View Image Challenge - Classification (SAR Classification) task, our model achieved an accuracy of 21.45\%, an AUC of 0.56, and a total score of 0.30, placing us 9th in the competition.
\end{abstract}

\section{Introduction}
\label{sec:intro}
Multimodal learning~\cite{ramachandram2017deep} is a research field that combines multiple sensors and data sources to enhance recognition, analysis, and learning capabilities. Recently, there has been growing interest in developing models and methods to improve the performance of Automatic Target Recognition (ATR)~\cite{ozkaya2020automatic} systems. In the context of the Multi-modal Aerial View Imagery Challenge - Classification, a sub-challenge of the PBVS 2024 workshop~\cite{low2024multi}, participants are tasked with identifying ten different types of vehicles using Synthetic Aperture Radar (SAR) data, which is robust against weather and visible light conditions. Electro-Optical (EO) image data is also provided but only for training purposes.

\begin{figure}[t!]  
  \centering  
  \includegraphics[width=1.0\linewidth]{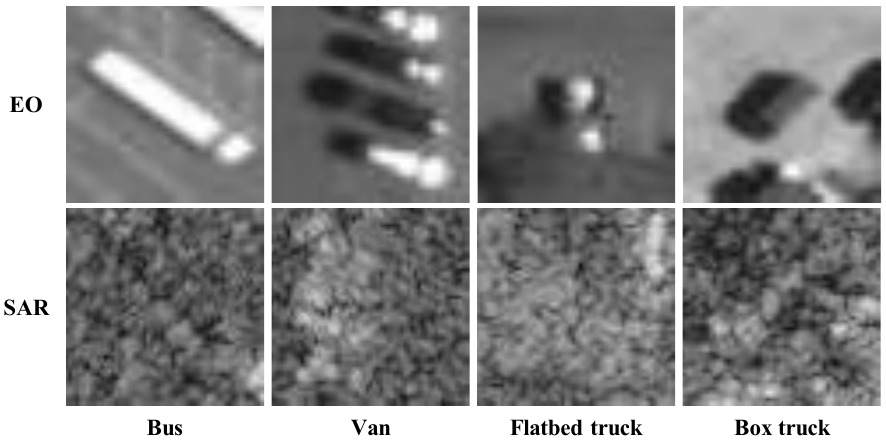}  
  \caption{Provided SAR-EO Pair Dataset.}  
  \label{fig:SAR-EO}  
  \vspace*{-0.3cm}  
\end{figure}

The dataset features a severe long-tail distribution, with the largest class containing 364,291 samples, while the smallest class has only 353 samples—a disparity exceeding 1,000 times. Furthermore, the EO and SAR images vary in size, with EO images typically measuring 31x31 pixels and SAR images 51x51 pixels. SAR images, in particular, are noisy and difficult to interpret, adding complexity to the task. Figure~\ref{fig:SAR-EO} shows examples of provided SAR-EO pair dataset.

\begin{figure*}[t!]
  \centering
  \includegraphics[width=1.0\linewidth]{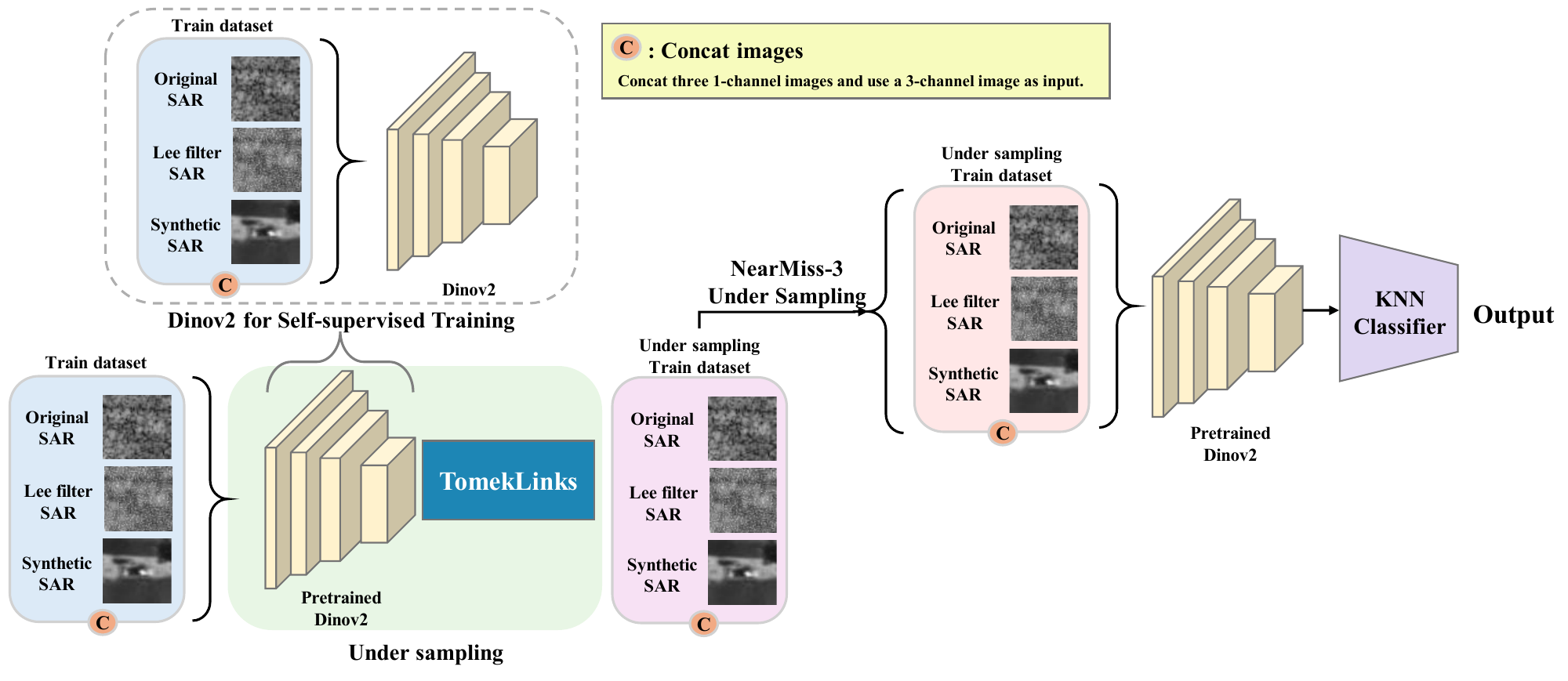}
  \caption{Overview of Our Proposed Pipeline}
  \label{fig:pipeline}
  \vspace*{-0.3cm}
\end{figure*}

In this study, we aim to address two critical challenges to improve recognition performance:  

Firstly, the \textbf{severe long-tail distribution} of the dataset. Recent research~\cite{samuel2021distributional, zhang2021distribution} on long-tail distributions suggests two-stage or multi-stage training approaches. In the first stage, the entire dataset is used to train the feature extractor, which is then frozen, and the classifier is trained in the second stage using sampling techniques or loss functions to create a more balanced classifier. While these approaches are effective for imbalances of 10x to 100x, they struggle with datasets exhibiting 1,000x imbalance, as the initial feature extractor is biased towards head classes, influencing the second-stage classifier to overfit the head class. Furthermore, uniform sampling often leaves a large portion of the data underutilized, reducing generalization performance.  

Secondly, the \textbf{domain disparity} between SAR and EO images. Traditional multimodal methods~\cite{zhou2023cmot, zhang2023cmx, kim2024privacy, hinton2015distilling} predominantly rely on knowledge distillation techniques based on features or logits. However, differences in resolution and the inconsistency of information between SAR and EO images, coupled with significant noise in SAR data, make such methods less effective.

\noindent To address these challenges, we propose a novel approach. An overview of our proposed pipeline can be seen in Figure~\ref{fig:pipeline}.
\begin{itemize}
  \item To tackle \textbf{long-tail imbalance}, we combine \textbf{clustering} and \textbf{self-supervised learning} to reduce prediction bias. Clustering helps identify representative samples, while self-supervised methods provide additional training signals without requiring labeled data.
  \item To address \textbf{domain disparity}, we introduce noise-filtering techniques for SAR images and \textbf{SAR-to-EO translation}, which allows the model to use converted EO images for further analysis.  
\end{itemize}

\section{Methodology}
\label{sec:me}

\subsection{Data Preparation}  
\begin{figure}[t!]  
  \centering  
  \includegraphics[width=1.0\linewidth]{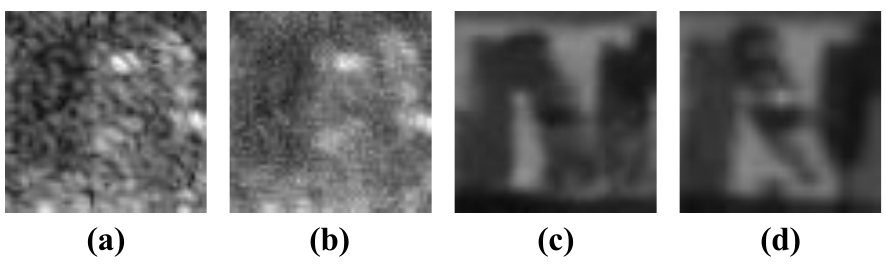}  
  \caption{\textbf{Types of Data Used in Training.} (a) is the original SAR image. (b) The SAR image with a Lee filter applied. (c) The result of translating the SAR image to EO using the Pix2PixHD model, and (d) The original EO image for comparison with (c).}  
  \label{fig:Synthetic}  
  \vspace*{-0.3cm}  
\end{figure}  

It is well-established that denoising SAR images improves classifier performance~\cite{yommy2015sar}. In this study, we adopt this strategy by applying the Lee filter denoising algorithm to SAR images. This filter effectively reduces noise while preserving the structural details of the images, enhancing classifier performance.  

To further improve accuracy and robustness by leveraging the complementary information between SAR and EO imagery, we perform SAR-to-EO translation~\cite{cabrera2021sar, seo2023improved, kim2024clean} using the Pix2PixHD model~\cite{wang2018high}. However, both denoising and image-to-image translation can occasionally distort shape information, especially for uncommon distributions. To mitigate this, we concatenate the original SAR image, the denoised SAR image, and the SAR-translated EO image into a three-channel input, which is used for both training and inference. Figure~\ref{fig:Synthetic} shows examples of the original, denoised, and SAR-to-EO translated images.  

\subsection{SAR Classification}  
To address the long-tailed distribution, we adopt a two-step training process based on self-supervised learning with the DINOv2 model.  

\textbf{Step 1:} Self-Supervised Training. The DINOv2 model~\cite{oquab2023dinov2} is trained on the entire dataset in a self-supervised manner to extract robust features without requiring labeled data. This training enables the model to learn robust representations without requiring labeled data.

\textbf{Step 2:} Data Balancing and Classifier Training. After training, the DINOv2 model extracts high-dimensional features from all images. We apply the Tomek Links method to remove majority-class samples near minority-class boundaries and use NearMiss-3 sampling to balance ambiguous decision boundaries. This results in N balanced subsets containing balanced samples from each category, significantly reducing data imbalance. Each subset is used to train a K-Nearest Neighbors (KNN) classifier. Finally, the predictions of the N classifiers are combined using an ensemble technique, which improves both accuracy and generalization performance. 

By effectively addressing the category imbalance present in long-tailed datasets, this method significantly improves object recognition performance, especially for rare categories.

\section{Experiments}
We trained our model on approximately 455,600 SAR-EO pairs provided in the challenge dataset, applying our comprehensive three-channel preprocessing pipeline to enhance the input data. This preprocessing step involved concatenating three distinct types of SAR-related data: the original SAR images, denoised SAR images processed using the Lee filter algorithm to reduce noise while preserving structural details, and synthetic SAR-to-EO translated images generated using the Pix2PixHD model~\cite{wang2018high}. By integrating these complementary data types, we created a rich and informative input representation for both training and inference.

For model training, we employed the DINOv2~\cite{oquab2023dinov2} framework, a state-of-the-art self-supervised learning model based on the Vision Transformer (ViT)~\cite{dosovitskiy2020image} architecture. We initialized the model with pre-trained weights from DINOv2 to leverage its robust feature extraction capabilities. The input images were resized to 56x56 pixels to ensure consistent dimensions for processing. The ensemble consisted of 7 subsets, each using K=3 neighbors. 

This multi-step approach not only mitigated the effects of class imbalance but also enhanced the model's generalization across rare and underrepresented categories.

As a result, our methodology achieved an accuracy of 21.45\% and an AUC of 0.56, yielding a total score of 0.30 and securing 9th place in the overall rankings. The final performance and rankings of the top 10 teams are summarized in Table ~\ref{table:rank}.

\begin{table}[htbp]
    \centering
    \small
    \caption{Top-10 Teams for MAVIC-C.}
    \label{table:rank}
    \begin{tabular}{ccccc}
        \toprule
        Rank & Team & Total Score & Accuracy & AUC \\
        \midrule
        1    & IQSKJSP   & 0.49        & 37.9     & 0.83 \\
        2    & MITHF     & 0.46        & 38.85    & 0.69 \\
        3    & GuanYu    & 0.39        & 35.10    & 0.49 \\
        4    & GWLOong   & 0.35        & 38.80    & 0.24 \\
        5    & http      & 0.33        & 10.40    & 1.00 \\
        6    & unknown   & 0.33        & 10.05    & 1.00 \\
        7    & findanswear & 0.32      & 10.00    & 1.00 \\
        8    & NJUST-KMG & 0.31        & 8.45     & 1.00 \\
        9    & yuhyun    & 0.30        & 21.45    & 0.56 \\
        10   & bingxiang  & 0.30       & 18.90    & 0.62 \\
        \bottomrule
    \end{tabular}
\end{table}

\section{Conclusion}

This study introduces comprehensive methods to enhance the classification performance of SAR images by leveraging multimodal SAR-EO data, which combines the strengths of both modalities. SAR data, while robust to weather conditions and independent of visible light, is inherently noisy and challenging to interpret. To address these limitations, we developed a novel preprocessing pipeline that integrates denoised SAR data, SAR-to-EO synthetic imagery, and the original SAR data into a unified input. This approach provides richer and more diverse information for the classification model, enabling better exploitation of the complementary characteristics of SAR and EO data.

A key innovation in our methodology is the use of a self-supervised learning approach paired with a data balancing process to tackle the severe long-tailed distribution present in the dataset. By training the DINOv2 model, a state-of-the-art self-supervised framework, we were able to effectively capture robust and generalizable features from the data without relying on extensive labeled samples. The Tomek Links and NearMiss-3 techniques were instrumental in refining the training dataset by balancing class distributions and focusing on ambiguous decision boundaries, particularly for rare categories. These balanced subsets provided an optimal foundation for training the K-Nearest Neighbors (KNN) classifiers.

Additionally, our ensemble strategy, which aggregates the predictions of multiple KNN classifiers trained on balanced subsets, further boosts the robustness and accuracy of the model. By mitigating the limitations of individual classifiers, the ensemble approach enhances performance across all categories, with particular gains observed in underrepresented and rare classes.

In conclusion, this study demonstrates innovative techniques that combine multimodal data, self-supervised learning, and ensemble methods to address key challenges in SAR image classification, including noise, long-tailed distributions, and limited interpretability. These methodologies provide significant insights into SAR-based object recognition and hold substantial potential for advancing future research and technological developments in the field.


{\small
\bibliographystyle{ieee_fullname}

\bibliography{egbib}
}

\end{document}